\documentclass[conference]{IEEEtran}
\IEEEoverridecommandlockouts
\usepackage{cite}
\usepackage{amsmath,amssymb,amsfonts}
\usepackage{algorithmic}
\usepackage{algorithm}
\usepackage{graphicx}
\usepackage{textcomp}
\usepackage{bm}
\usepackage{booktabs}
\usepackage{hyperref}
\usepackage{url}
\usepackage{xcolor}
\usepackage{amsmath}
\usepackage{multirow}
\usepackage{color}
\usepackage{graphicx}
\def\BibTeX{{\rm B\kern-.05em{\sc i\kern-.025em b}\kern-.08em
    T\kern-.1667em\lower.7ex\hbox{E}\kern-.125emX}}
\begin{document}

\newcommand{\hong}[1]{{\color{black} #1}\normalfont}

\title{Distributed Pruning Towards Tiny Neural Networks in Federated Learning\\

}


\author{Hong Huang$^1$, Lan Zhang$^2$, Chaoyue Sun$^3$, Ruogu Fang$^3$, Xiaoyong Yuan$^2$, Dapeng Wu$^1$\\
$^1$City University of Hong Kong, $^2$Michigan Technological University, $^3$University of Florida\\
\texttt{honghuang2000@outlook.com, lanzhang@mtu.edu, chaoyue.sun@ufl.edu}\\ \texttt{ruogu.fang@bme.ufl.edu, xyyuan@mtu.edu, dpwu@ieee.org}
}

\maketitle

\begin{abstract}
Neural network pruning is an essential technique for reducing the size and complexity of deep neural networks, enabling large-scale models on devices with limited resources. However, existing pruning approaches heavily rely on training data for guiding the pruning strategies, making them ineffective for federated learning over distributed and confidential datasets. Additionally, the memory- and computation-intensive pruning process becomes infeasible for recourse-constrained devices in federated learning. To address these challenges, we propose FedTiny, a distributed pruning framework for federated learning that generates specialized tiny models for memory- and computing-constrained devices. We introduce two key modules in FedTiny to adaptively search coarse- and finer-pruned specialized models to fit deployment scenarios with sparse and cheap local computation. First, an adaptive batch normalization selection module is designed to mitigate biases in pruning caused by the heterogeneity of local data. Second, a lightweight progressive pruning module aims to finer prune the models under strict memory and computational budgets, allowing the pruning policy for each layer to be gradually determined rather than evaluating the overall model structure.
The experimental results demonstrate the effectiveness of FedTiny, which outperforms state-of-the-art approaches, particularly when compressing deep models to extremely sparse tiny models. FedTiny achieves an accuracy improvement of 2.61\% while significantly reducing the computational cost by 95.91\% and the memory footprint by 94.01\% compared to state-of-the-art methods.
\end{abstract}

\section{Introduction}
Deep neural networks (DNNs) have achieved great success in the past decade. However, the huge computational cost and storage overhead limit the usage of DNNs on resource-constrained devices. Neural network pruning has been a well-known solution to improve hardware efficiency~\cite{janowsky1989pruning, han2015deep}. The core of neural network pruning is to remove insignificant parameters from a DNN and determine specialized subnetworks for different hardware platforms and training tasks (defined as deployment scenarios). 
To achieve better accuracy, most pruning approaches rely heavily on training data to trade off model size, efficiency, and accuracy~\cite{han2015deep,louizos2018learning, yu2018nisp, molchanov2019importance,singh2020woodfisher}, which, unfortunately, becomes ineffective when dealing with confidential training datasets distributed over resource-constrained devices.

\begin{figure*}[!t]
\centering
\includegraphics[width=0.85\textwidth]{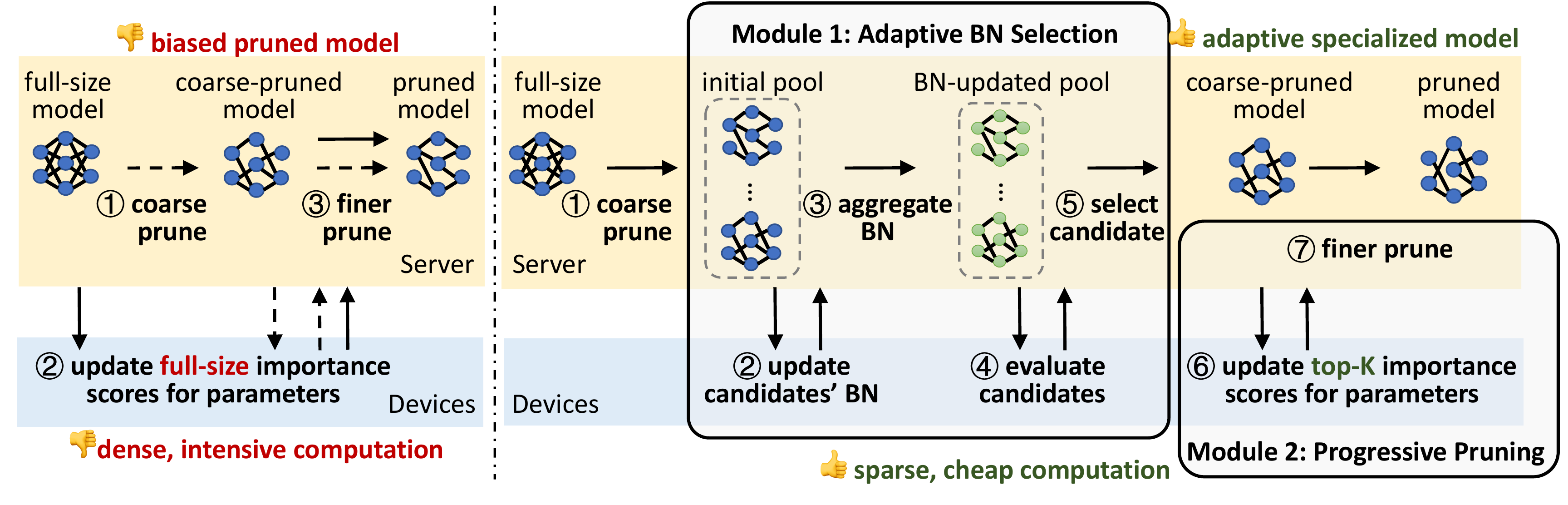} 
\caption{Overview of FedTiny for the specialized tiny model in federated learning. \textit{Left}: Existing federated pruning approaches push pruning operations to devices. Either a full-size model (solid arrow) or a coarse-pruned model (dash arrow) is finer-pruned under dense and intensive local computation, suffering biased pruning. \textit{Right}: FedTiny introduces two key modules, the adaptive batch normalization module and the progressive pruning module, to adaptively search coarse- and finer-pruned specialized models to fit deployment scenarios with sparse and cheap local computation.
}
\label{fig:framework}
\end{figure*}

Recent success in federated learning enables collaborative training across distributed devices with confidential local datasets~\cite{li2020federated}. Instead of uploading local data, federated learning aggregates on-device knowledge by iteratively updating local model parameters at the server. While successful, federated learning cannot determine the specialized pruned model for participating devices without training data. To address this issue, \cite{xu2021accelerating} proposed to decouple the pruning process under federated environments, where a large-size model is first pruned on the server and then fine-tuned on devices.
However, since most pruning algorithms require a guide from the data distribution, without access to device-side training data, the server-side pruning leads to significant bias in the pruned subnetwork, especially under heterogeneous (non-iid) local data distributions.
To mitigate such bias issues, recent research pushes pruning operations to devices~\cite{shao2019privacy, li2021lotteryfl, munir2021fedprune, liu2021adaptive, jiang2022model}. As shown in Fig.~\ref{fig:framework} left, either a full-size model or a coarse-pruned model will be finer-pruned based on the updated importance scores from devices. The importance scores for all parameters need to store in memory, which is infeasible for resource-constrained devices with limited memory budgets. Moreover, without any interaction with the device side, the initial model through server-side coarse pruning still suffers from the bias issue, requiring extra efforts in later finer pruning to find the optimal subnetwork. Such negative impact becomes more challenging when pruning towards an extremely tiny subnetwork, as the biased initial subnetwork can deviate significantly from the optimal structure, resulting in poor accuracy~\cite{evci2020rigging}.

To address the above challenges, in this paper, we develop a novel distributed pruning framework for federated learning named FedTiny. Depending on the deployment scenarios, \textit{i.e.}, participating hardware platforms, and training tasks, FedTiny can obtain specialized tiny models using distributed and confidential datasets on participating devices. Besides, FedTiny allows devices with tight memory and computational budgets to participate in the resource-intensive pruning process by reconfiguring interactions between the server and devices. As shown in Fig.~\ref{fig:framework} right, FedTiny introduces two key modules: the adaptive batch normalization (BN) selection module and the progressive pruning module. 
To avoid the negative impact of biased initial pruning, we introduce the \textit{adaptive BN module} to identify a specialized coarse-pruned model by indirectly pruning at devices, where devices only evaluate the server-side pruning. It should be mentioned that evaluating a pruned model is much cheaper than training and pruning. The local evaluation is feedback to the server through batch normalization parameters. Since batch normalization layers can effectively measure local data distribution with very few parameters~\cite{ioffe2015batch}, this module guides the initial pruning with little computation and communication cost. 
Besides, contrary to prior research using importance scores of all parameters in a full-size model for finer pruning, the \textit{progressive pruning module} is developed to iteratively adjust the model structure with sparse and cheap local computation. Inspired by RigL~\cite{evci2020rigging}, devices only rate partial model parameters (\textit{e.g.}, a single layer) at a time, where the top-K importance scores are stored locally and uploaded to the server, significantly reducing memory, computation, and communication cost. 

To demonstrate the effectiveness of FedTiny, we evaluate FedTiny on ResNet18~\cite{he2016deep} and VGG11~\cite{simonyan2014very} with four image classification datasets (CIFAR-10, CIFAR-100, CINIC-10, and SVHN). Extensive experimental results suggest that FedTiny achieves much higher accuracy with a lower level of memory and computational cost than state-of-the-art baseline approaches. Especially in a low-density regime~\cite{hoefler2021sparsity} from $10^{-2}$ to $10^{-3}$, FedTiny gets a slight loss of accuracy, while other baselines suffer from the sharp drop in accuracy. Moreover, FedTiny achieves top-one accuracy of $85.23\%$ with the $0.014\times$ FLOPs and $0.03\times$ memory footprint of ResNet18~\cite{he2016deep}, which outperforms the best baseline, which gets $82.62\%$ accuracy with $0.34 \times$ FLOPs and $0.51\times$ memory footprint.

\section{Related Work}


\subsection{Neural Network Pruning} 
Neural network pruning has been a well-known technique to remove redundant parameters of a DNN for model compression, which can trace back to the late 1980s~\cite{mozer1988skeletonization, lecun1989optimal, janowsky1989pruning}. Most existing pruning approaches focus on the trade-off between accuracy and sparsity in the \textit{inference} stage. A typical pruning process first calculates the importance scores of all parameters in a well-trained DNN and then removes parameters with lower scores. The importance scores can be derived based on the weight magnitudes~\cite{janowsky1989pruning, han2015deep}, the first-order Taylor expansion of the loss function~\cite{mozer1988skeletonization, molchanov2019pruning}, the second-order Taylor expansion of the loss function~\cite{lecun1989optimal, molchanov2019importance}, and other variants~\cite{louizos2018learning, yu2018nisp, singh2020woodfisher}. 

Another line of recent research on neural network pruning focuses on improving the efficiency of the \textit{training} stage, which can be divided into two categories. One is pruning at initialization, \textit{i.e.}, pruning the original full-size model before training. The pruning policy can be determined by evaluating the connection sensitivity~\cite{lee2018snip}, Hessian-gradient product~\cite{wang2020picking}, and synaptic flow~\cite{tanaka2020pruning} of the original model. Since such pruning does not involve the training data, the pruned model is not specialized for the training task, resulting in biased performance. The other category is dynamic sparse training~\cite{mocanu2018scalable,dettmers2019sparse, evci2020rigging}. The pruned model structure is iteratively adjusted throughout the training process while maintaining the pruned model size at the desired sparsity. 
However, the pruning process is to adjust the model structure in a large search space, requiring memory-intensive operations, which is infeasible for resource-constrained devices. Although RigL~\cite{evci2020rigging} tries to reduce memory consumption, it needs to compute gradients for all parameters, which is computationally expensive and may lead to straggling issues in federated learning.
\subsection{Neural Network Pruning in Federated Learning} 
Federated Learning has recently gained attention as a promising approach to address data privacy concerns in collaborative machine learning. FedAvg~\cite{mcmahan2017communication}, one of the most widely used methods in federated learning, utilizes locally updated on-device models instead of raw data to achieve private knowledge transferring.
Since data is locally stored and cannot be shared, the aforementioned pruning approaches that rely on training data cannot be used in federated learning. Enlighten by pruning at initialization, Xu et al. proposed to prune the original full-size model at the server and fine-tune at devices with their local data~\cite{xu2021accelerating}. Existing pruning at initialization approaches, such as SNIP~\cite{lee2018snip}, GraSP~\cite{wang2020picking}, and SynFlow~\cite{tanaka2020pruning}, can be directly converted to server-side pruning. However, server-side pruning usually results in significantly biased pruned models, especially for heterogeneous (non-iid) local data distributions.

To mitigate such bias, recent research pushes pruning operations under federated settings to devices. By locally training a full-size model, SCBF~\cite{shao2019privacy} dynamically discards the unimportant channels on devices. Such local training with a full-size model is assigned to a part of devices in FedPrune to guide pruning based on the updated activations~\cite{munir2021fedprune}. 
Besides, LotteryFL~\cite{li2021lotteryfl} iteratively prunes a full-size model on devices with a fixed pruning rate to find a personalized local subnetwork. However, the above research suffers from large memory and computational cost on the device side because devices need to locally compute the importance scores of all parameters. Although PruneFL~\cite{jiang2022model} reduces the local computational cost by finer pruning a coarse-pruned model rather than a full-size model, it still requires a large local memory footprint to record the updated importance scores of all parameters in the full-size model. 
ZeroFL~\cite{qiu2022zerofl} partitions weights into active weights and non-active weights in the inference and sparsified weights and activations for backward propagation. However, this approach still needs a large memory space because the non-active weights and the gradients generated through the training process are still stored in a dense fashion. 
FedDST~\cite{bibikar2022federated} deploys the mask adjustment on the devices, and the server generates a new global model via sparse aggregation and magnitude pruning. It needs much more computation cost because it needs extra training epochs to recover the growing weights before uploading, which may lead to straggling issues in federated learning. The coarse-pruned model still suffers from bias issues in the server-side pruning.
Existing federated neural network pruning fails to obtain a specialized tiny model without bias and memory-/compute-budget concerns. Therefore, we develop FedTiny to achieve this.

\subsection{Federated Learning With Non-iid Data}

Federated learning suffers from divergence when the data distributions across devices are heterogeneous (non-iid)~\cite{zhao2018federated}. Several works have been proposed to address non-iid challenges, \textit{e.g.}, MATCHA~\cite{marfoq2020throughput}, FedProx\cite{li2020federated}, and FedNova\cite{wang2020tackling}. 
These works provide the convergence guarantees of federated learning under strong assumptions, which becomes impractical in real-world scenarios.

Data augmentation (\textit{e.g.}, Astraea~\cite{duan2019astraea}, FedGS~\cite{li2022data}, and CSFedAvg~\cite{zhang2021client}) and personalization methods (\textit{e.g.}, meta learning~\cite{chen2018federated}, multi-task learning~\cite{smith2017federated}, and knowledge distillation~\cite{lin2020ensemble}) are two promising approaches to address non-iid issues. However, these methods are computationally intensive and become infeasible in resource-constrained scenarios. 
In our work, we develop a novel distributed pruning approach with adaptive batch normalization selection to find an unbiased coarse-pruned model for addressing the non-iid challenges in resource-constrained devices.
\section{Proposed FedTiny}
This section introduces the proposed FedTiny. We first describe the problem statement, followed by our design principles. Accordingly, we present two key modules in FedTiny: the adaptive BN selection module and the progressive pruning module.

\subsection{Problem Statement}
We consider a typical federated learning setting, where $K$ devices collaboratively train a neural network with their corresponding local datasets $\mathcal{D}_k, k \in \{1,2,\dots, K\}$. All devices have limited memory and computing resources.
Given a large neural network with dense parameters $\bm{\Theta}$, we aim to find a specialized subnetwork with sparse parameters $\bm{\theta}$ and mask $\bm{m}$ on dense parameters to achieve the optimal prediction performance for federated learning. The sparse parameters are derived by applying a mask to the dense parameters: $\bm{\theta} = \bm{\Theta} \odot \bm{m}$ ($\bm{m} \in \{0, 1\}^{|\bm{\Theta}|}$). During training, density $d$ of sparse mask $\bm{m}$ cannot exceed target density $d_{target}$. $d_{target}$ is determined by the limitation of devices' memory resources. We formulate the problem as a constrained optimization problem:
\begin{equation}
    \begin{aligned}
    \min_{\bm{\theta}, m}  \quad & \sum_{k=1}^{K} L(\bm{\theta}, \bm{m},\mathcal{D}_k), \\
    \textrm{s.t.} \quad & d \le d_{target}
    \end{aligned}
    \label{eq:obj}
\end{equation}
where $L(\bm{\theta}, \bm{m},\mathcal{D}_k)$ denotes the loss function for local dataset $\mathcal{D}_k$ on the $k$-th device.

\subsection{Design Principles}
As shown in Fig.~\ref{fig:framework} left, existing federated neural network pruning faces two main challenges, bias in coarse pruning and intensive memory consumption in finer pruning. To address these challenges, we propose a FedTiny.
The overview of FedTiny is illustrated in Fig.~\ref{fig:framework} right, which consists of two key modules: the adaptive BN selection module and the progressive pruning module.

The adaptive batch normalization selection module (Steps 2-5 in Fig.~\ref{fig:framework} right)  aims to derive an adaptive coarse-pruned structure on the server and alleviate bias in the coarse pruning due to unseen heterogeneous data over devices. In this module, devices first collaboratively update batch normalization measurements for all candidate models from coarse pruning. Then the server selects one less biased candidate model as the initial coarse-pruned model based on device evaluations.

The progressive pruning module (Steps 6-7 in Fig.~\ref{fig:framework} right) further improves the coarse-pruned model by finer pruning at resource-constrained devices, significantly reducing the on-device memory footprint and computational cost. In this module, the devices only maintain the top-K importance scores of the pruned parameters. Based on the average importance scores, the server grows and prunes parameters to produce a new model structure. After iterative growing and pruning, the model structure progressively approaches the optimal structure.

In the following, we provide detailed descriptions of the adaptive batch normalization selection module and the progressive pruning module, respectively.

\subsection{Adaptive Batch Normalization Selection}
\label{sec:ABNS}
It is critical to address the bias issue in the coarse-pruned model, as the highly biased pruned structure requires more resources and time to adjust to the optimal structure, especially in the low-density regime.
One possible approach is to send a set of pruned structure candidates to the devices and let devices select the least biased model from the candidate pool. We call this approach vanilla selection~\cite{he2018amc}. 
However, recent research~\cite{li2020eagleeye} shows that pruned model performance varies before and after fine-tuning, which makes the pruned structure candidate selected before fine-tuning not necessarily the best one after fine-tuning. 
Such an issue could be exaggerated in the federated settings as the heterogeneous data distribution over devices may further increase the discrepancy of pruned model performance in fine-tuning. 


To address this issue, we introduce adaptive batch normalization selection in FedTiny. Adaptive batch normalization selection updates batch normalization measurements for candidate models before evaluation, aiming to derive a less biased coarse-pruned structure. The algorithm of the adaptive batch normalization selection module is illustrated in Algorithm~\ref{alg:abns}.

We introduce batch normalization (BN)~\cite{ioffe2015batch} to provide measurements for data distribution across devices. Such measurements provide representations of on-device data and thus guide the pruning process. The batch normalization transformation is calculated upon the following transformation on $i$-th input $x_i$ in each batch,
\begin{equation}
\hat{x}_i \leftarrow \frac{x_i - \mu}{\sqrt{\sigma^2 + \epsilon}},
\end{equation}
where $\epsilon$ is a small constant. During training, $\mu$ and $\sigma$ are updated based on moving mean $\mu_i$ and standard deviation $\sigma_i$ of the batch $x_i$, 
\begin{equation}
\mu_t = \gamma\mu_{t-1} + (1 - \gamma)\mu_i, \quad \sigma^2_t = \gamma\sigma^2_{t-1} + (1 - \gamma)\sigma^2_i,
\end{equation}
where $\gamma$ denotes the momentum coeffcient and $t$ is the number of training iterations. During testing, the mean $\mu$ and standard deviation $\sigma$ keep fixed.

In the adaptive batch normalization selection module, batch normalization measurements are updated in the forward pass on devices before evaluation to select a less biased-coarse pruned candidate.
Specifically, after coarse pruning on full-size parameters $\bm{\Theta}$ with different strategies, the server obtains an initial pool consisting of $C$ candidate models with their sparse parameters $\bm{\theta}^{(c)}$ and the corresponding masks $\bm{m}^{(c)}$, where $\bm{\theta}^{(c)} = \bm{\Theta} \odot \bm{m}^{(c)}$, for $c \in \{1, 2, \dots, C\}$. For each candidate model, we set different pruning ratios for each layer while keeping overall density $d \leq d_{target}$. 
Devices first fetch all candidate models. Note that the communication cost is low due to the ultra-low network density.
Then, each device (say the $k$-th) samples a development dataset from local data, $\hat{\mathcal{D}}_k \subset \mathcal{D}_k$, freezes all parameters and updates the means $\mu_k^{(c)}$ and standard deviations $\sigma_k^{(c)}$ of batch normalization layers in the $c$-th candidate model.
Next, the server aggregates all local batch normalization measurements from devices to obtain new global batch normalization measurements for each candidate model, \textit{i.e.}, for $c \in \{1,2,\dots, C\}$,
\begin{equation}
    \mu^{(c)} = \sum_{k=1}^{K}\frac{|\hat{\mathcal{D}}_k|}{\sum_{k=1}^{K} |\hat{\mathcal{D}}_k|} \mu^{(c)}_k ,\quad \sigma^{(c)} = \sum_{k=1}^{K}\frac{|\hat{\mathcal{D}}_k|}{\sum_{k=1}^{K} |\hat{\mathcal{D}}_k|} \sigma^{(c)}_k,
    \label{eq:aggbn}
\end{equation}
where $|\hat{\mathcal{D}}_k|$ denotes the number of samples in the dataset $\hat{\mathcal{D}}_k$. 

\begin{algorithm}[H]
\caption{Adaptive batch normalization selection}
\label{alg:abns}
\textbf{Input}: $C$ coarse-pruned candidate models with sparse parameters $\bm{\theta}^{(1)}, \dots, \bm{\theta}^{(C)} $ and their corresponding masks $\bm{m}^{(1)}, \dots, \bm{m}^{(C)}$  on the server, $K$ devices with local development dataset $ \hat{\mathcal{D}}_1, \dots, \hat{\mathcal{D}}_K $.\\
\textbf{Output}: the less biased coarse-pruned model with parameters $\bm{\theta}_0$  and its corresponding mask $\bm{m}_0$.\\
\begin{algorithmic}[1]
\STATE // Device-side 
 \FOR{$k = 1 $ to $K$}
\STATE Fetch sparse parameters $\bm{\theta}^{(1)}, \bm{\theta}^{(2)}, \dots, \bm{\theta}^{(C)}$ and their corresponding masks $\bm{m}^{(1)},\bm{m}^{(2)}, \dots, \bm{m}^{(C)}$ from the server
 \FOR{$c = 1$ to $C$}
  \STATE Calculate local batch normalization measurements $\mu_k^{(c)}$, $\sigma_k^{(c)}$ by forward pass on $\bm{\theta}^{(c)}$ with dataset $\hat{\mathcal{D}}_k$
 \ENDFOR
\STATE Upload $\mu_k^{(1)},\mu_k^{(2)}, \dots, \mu_k^{(C)}$ and $\sigma_k^{(1)},\sigma_k^{(2)}, \dots, \sigma_k^{(C)}$ to the server
\ENDFOR
\STATE // Server-side 
\FOR{$c = 1$ to $C$} 
\STATE $\mu^{(c)} = \sum_{k=1}^{K}\frac{|\hat{\mathcal{D}}_k|}{\sum_{k=1}^{K} |\hat{\mathcal{D}}_k|} \mu^{(c)}_k$
\STATE $\sigma^{(c)} = \sum_{k=1}^{K}\frac{|\hat{\mathcal{D}}_k|}{\sum_{k=1}^{K} |\hat{\mathcal{D}}_k|} \sigma^{(c)}_k$
\ENDFOR
\STATE // Device-size
\FOR{$k = 1 $ to $K$}
\STATE Fetch ${\mu}^{(1)}, {\mu}^{(2)}, \dots, {\mu}^{(C)}$ and ${\sigma}^{(1)},{\sigma}^{(2)}, \dots, {\sigma}^{(C)}$ from the server
\FOR{$c = 1$ to $C$ }
\STATE $\mu_{k}^{(c)}, \sigma_{k}^{(c)} \leftarrow \mu^{(c)}, \sigma^{(c)}$ // Each candidate model installs global batch normalization measurements
\STATE $s^{(c)}_k \leftarrow L(\bm{\theta}^{(c)},\bm{m}^{(c)}, \hat{\mathcal{D}}_k)$ // Calculate the loss as evaluation metrics
\ENDFOR
\STATE Upload $s^{(1)}_k, s^{(2)}_k, \dots,s^{(C)}_k$ to the server
\ENDFOR
\STATE // Server-side
\FOR{$c = 1$ to $C$}
\STATE $s^{(c)} \leftarrow \sum_{k=1}^K\frac{|\hat{\mathcal{D}}_k|}{\sum_{k=1}^{K} |\hat{\mathcal{D}}_k|} s^{(c)}_k$ 
\ENDFOR
\STATE $c^* \leftarrow \mathrm{argmin}_c({s^{(c)}})$ // Select the candidate model with the lowest loss
\STATE $\bm{\theta}_0, \bm{m}_0 = \bm{\theta}^{(c^*)}, \bm{m}^{(c^*)}$
\RETURN $\bm{\theta}_0, \bm{m}_0$
\end{algorithmic}
\end{algorithm}

After that, each device updates global batch normalization measurements $\mu^{(c)}, \sigma^{(c)}$ for $c$-th candidate model. Considering Eq.~\ref{eq:obj}, we let devices calculate evaluation loss for each updated candidate model with their on-device data and let the server select the candidate model with the lowest average loss as the coarse-pruned model. 

Note that although the adaptive batch normalization selection module requires transferring the parameters, the communication cost remains minimal, as only the parameters in pruned models with ultra-low density need to be transferred. The detailed analysis of communication costs is discussed in Section~\ref{sec:eval_impact}. Additionally, batch normalization transformation is calculated as part of the forward pass at the device without gradient calculation or updates. Therefore, the adaptive batch normalization selection effectively addresses the bias of model structure without incurring significant memory or computational overhead.

\subsection{Progressive Pruning}
\label{sec:pp}
Given a coarse-pruned model from the above module, we introduce progressive pruning to further fine-prune the model for better performance.
We propose the progressive pruning module with two improvements: 1) only the top-K importance scores are calculated, while the remaining importance scores are discarded to save memory space;
2) partial model parameters (\textit{e.g.}, a single layer) are adjusted per rounds instead of the entire model to avoid intensive computation.
FedTiny utilizes a growing-pruning adjustment on the model structure while maintaining the sparsity. 
Specifically, the server grows the pruned parameters and prunes the same number of unpruned parameters to adjust the model structure. Denote $a^l_t$ as the number of parameters that will be grown and pruned on layer $l$ at the $t$-th iteration. To guide growing and pruning on the server, each device only trains the sparse model and computes the Top-$a^l_t$ gradients for pruned parameters, which keeps the low memory footprint and computational cost in the resource-constrained device.
Furthermore, to reduce intensive computation, FedTiny divides the model structure into several blocks and prunes a block in one round. The progressive pruning module is detailed in Algorithm~\ref{alg:pp}.

In detail, each device (say the $k$-th) first downloads global sparse model parameter $\bm{\theta}_t$ with mask $\bm{m}_t$ as their local parameters $\bm{\theta}^k_{t}$ in the $t$-th iteration, and applies SGD with sparse gradients:
\begin{equation}
    \bm{\theta}^k_{t+1} = \bm{\theta}^k_{t} - \eta_{t}\nabla L(\bm{\theta}^k_{t}, \bm{m}_{t},\mathcal{B}_{t}^k) \odot \bm{m}_{t},
    \label{eq:localupdate}
\end{equation}
where $\eta_{t}$ is the learning rate, $\mathcal{B}_{t}^k$ is a batch of sample from the local dataset $\mathcal{D}_k$, and $\nabla L\odot \bm{m}_{t}$ denotes the sparse gradients for the sparse parameter $\bm{\theta}^k_{t}$.
After $E$ iterations of local SGD, each device calculates the top-$a^l_t$ gradients for pruned parameters on each layer $l$ with a batch of samples. We denote $\tilde{\bm{g}}^{k,l}_t$ as the top-$a^l_t$ gradients of pruned parameter with the largest magnitude on $k$-th device:
\begin{equation}
     \tilde{\bm{g}}^{k,l}_t =  \mathrm{TopK} \left( \bm{g}^{k,l}_t, a^l_t \right),
  \label{eq:grow}
\end{equation}
where $\mathrm{TopK}(\bm{v}, k)$ is threshold function, the elements of $\bm{v}$ whose absolute value is less than the $k$-th largest absolute value are replaced with 0, and ${\bm{g}}^{k,l}_t$ is the gradients of pruned parameters on layer $l$.

\begin{algorithm}[H]
\caption{Progressive pruning}
\label{alg:pp}
\textbf{Input}: initial coarse-pruned parameters $\bm{\theta}_0$ with mask $\bm{m}_0$, $K$ devices with local dataset $\mathcal{D}_1, \dots \mathcal{D}_K$, iteration number $t$, learning rate $\eta_t$, pruning number $a^l_t$ for each layer $l$, the number of local iterations per round $E$, the number of rounds between two pruning operation $\Delta R$, and the rounds at which to stop pruning $R_{stop}$. \\
\textbf{Output}: a well-trained model with sparse $\theta_{t}$ and adjusted mask $m_{t}$\\
\begin{algorithmic}[1] 
\STATE $t \leftarrow 0$
\WHILE{}
\STATE // Device-side
\FOR{$k = 1$ to $K$}
\STATE Fetch sparse parameters $\bm{\theta}_t$ and mask $\bm{m}_t$ from the server
\FOR{$i = 0$ to $E-1$}
\STATE $\bm{\theta}^k_{t+i+1} \leftarrow \bm{\theta}^k_{t+i} - \eta_{t+i}\nabla L(\bm{\theta}_{t+i}, \bm{m}_{t},\mathcal{B}_{t+i}^k) \odot \bm{m}_{t}$
\ENDFOR
\STATE Upload $\bm{\theta}_{t+E}$ to the server
\IF{$t \mod \Delta RE = 0$ and $t \leq ER_{stop}$}
\FOR{each layer $l$ in model}
\STATE Compute top-$a^l_t$ gradients $\tilde{\bm{g}}^{k,l}_t$ using Eq.~\ref{eq:grow} with a memory space of $O(a_t^l)$
\STATE Upload $\tilde{\bm{g}}^{k,l}_t$ to the server
\ENDFOR
\ENDIF
\ENDFOR
\STATE // Server-side
\STATE Compute the global parameters $\bm{\theta}_{t+E}$ by averaging the parameters from the devices
\IF{$t \mod \Delta RE = 0$ and $t \leq ER_{stop}$}
\FOR{each layer $l$ in model}
\STATE $\tilde{\bm{g}}^l_t \leftarrow \sum_{k=1}^{K}\frac{|\mathcal{D}_k|}{\sum_{k=1}^{K}|\mathcal{D}_k|} \tilde{\bm{g}}^{k,l}_t$
\STATE $\bm{I}^l_{grow} \leftarrow$ the $a^l_t$ pruned indices with the largest absolute value in $\tilde{\bm{g}}^l_t$
\STATE $\bm{I}^l_{drop} \leftarrow$ the $a^l_t$ unpruned indices with smallest weight magnitude in $\bm{\theta}_{t+E}$
\STATE Compute the new mask $\bm{m}_{t+E}^l$ by adjusting $\bm{m}_{t}^l$ based on $\bm{I}^l_{grow}$ and $\bm{I}^l_{drop}$
\ENDFOR
\STATE $\bm{\theta}_{t+E} \leftarrow \bm{\theta}_{t+E} \odot \bm{m}_{t+E}$ // Prune the model using the updated mask
\ELSE
\STATE$\bm{m}_{t+E} \leftarrow \bm{m}_t$
\ENDIF
\STATE $t \leftarrow t + E$
\ENDWHILE
\end{algorithmic}
\end{algorithm}

To calculate $\tilde{\bm{g}}^{k,l}_t$, devices create a buffer in the memory to store $a^l_t$ gradients. When a gradient is calculated, and the buffer is full, if its magnitude is larger than the smallest magnitude in the buffer, this gradient will be pushed into the buffer, and the gradient with the smallest magnitude will be discarded. Otherwise, this gradient will be discarded. In this manner, devices only need $O(a^l_t)$ memory space to store gradients.


Next, the server aggregates sparse parameters and gradients to get average parameters and average gradients $\tilde{\bm{g}}^{l}_t$ for each layer $l$,
\begin{equation}
    \tilde{\bm{g}}^{l}_t = \sum_{k=1}^K \frac{|\mathcal{D}_k|}{\sum_{k=1}^{K}|\mathcal{D}_k|}\tilde{\bm{g}}^{k,l}_t,
\end{equation}
where $|\mathcal{D}_k|$ denotes the number of samples in dataset $\mathcal{D}_k$. Then, the server grows $a^l_t$ pruned parameters with the largest averaged gradients magnitude on each layer $l$.
After that, the server prunes $a^l_t$ unpruned parameters (excluding the parameters just grown) with the smallest magnitude on each layer $l$.


\begin{figure*}[!tb]
    \centering
    \includegraphics[width=0.8\textwidth]{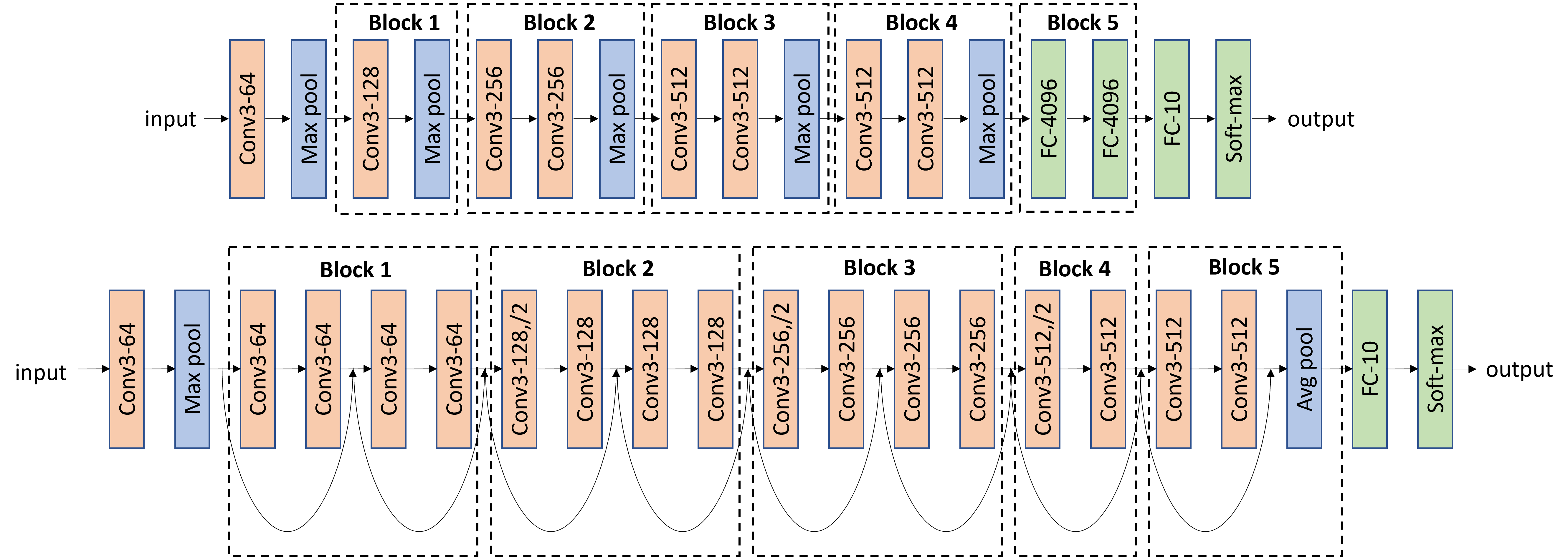}
    \caption{Partition of blocks of VGG11 (top) and ResNet18 (bottom) models.}
    \label{fig:grand}
\end{figure*}

According to growing and pruning, the server generates a global model with a new model structure, and FedTiny starts fine-tuning the new global model. FedTiny performs pruning and fine-tuning iteratively to achieve an optimal tiny neural network for all devices.
\section{Experiments}
In this section, we conduct comprehensive experiments on FedTiny. Firstly, we introduce the experiment setting and compare FedTiny with other baselines. Secondly, we conduct the ablation study to demonstrate the effectiveness of the adaptive batch normalization selection module and the progressive pruning module. Thirdly, we investigate the overhead in the adaptive batch normalization module and the impact of the pruning scheduling strategy. Fourthly, we demonstrate the effectiveness of FedTiny on heterogeneous data distributions. Finally, we compare the performance between FedTiny and small model training.

\subsection{Experimental Setup}
\label{sec:setup}
\subsubsection{Federated Learning Setting}
We evaluate FedTiny on image classification tasks with four datasets, CIFAR-10, CIFAR-100~\cite{krizhevsky2009learning}, CINIC-10~\cite{darlow2018cinic}, and SVHN~\cite{netzer2011reading} datasets on ResNet18~\cite{he2016deep} and VGG11~\cite{simonyan2014very} models. We consider $K=10$ devices in total.  For all datasets, we first generate various non-iid partitions on devices from Dirichlet distribution with $\alpha = 0.5$ and then change $\alpha$ in the Section~\ref{sec:non_iid}, following the setting in~\cite{luo2021no}. We train the models for 300 FL rounds on the CIFAR-10, CIFAR-100, and CINIC-10 datasets and 200 rounds on the SVHN dataset. Each round includes $5$ local epochs. The mini-batch size is set as 64. 

\subsubsection{FedTiny Setting}
We use the following settings in FedTiny.
In the adaptive batch normalization selection module, the server generates a candidate pool by magnitude pruning with various layer-wise pruning rate settings. Given target density, $d_{target}$, the server outputs candidates in the form of layer-wise pruning rate vectors $(d^1, d^2, \dots, d^L)$ for $L$-layer model based on Uniform Noise strategy. We derive the density $d^l$ for the $l$-th layer by adding the target density target with random noise $e^l$, \textit{i.e.}, $d^l =d_{target}+e^l$. A candidate can be added to the candidate pool only if its total density $d$ satisfies $d \le d_{target}$. After that, server can get a candidate pool $ \{\bm\theta^{(1)} , \bm\theta^{(2)},\dots, \bm\theta^{(C)} \}$ with mask $\{\bm{m}^{(1)}, \bm{m}^{(2)}, \dots, \bm{m}^{(C)} \}$. We first set the size of the candidate pool $C$ to 50 and then change the candidate pool size in Section~\ref{sec:eval_impact}. We set the ratio of the development dataset as 0.1, which is used to update local batch normalization measurements on the devices. In the progressive pruning module, we divide ResNet18 and VGG11 into five blocks and prune one block in each round, as shown in Fig.~\ref{fig:grand}. The order in which the server selects a block is backward, \textit{i.e.}, from the output layer to the input layer. We also evaluate pruning a single layer and pruning the entire model per round in Section~\ref{sec:impact_sch}.
The pruning number is set as $a^l_t = 0.15(1 + cos\frac{t\pi}{R_{stop}E})n^l$ for layer $l$ that will be pruned at the $t$-th iteration, where $n^l$ is the number of unpruned parameters in $l$-th layer. For layer $l$ that will not be pruned in the $t$-th iteration, $a^l_t=0$. We do not prune the batch normalization layer, bias, input layer, and output layer because they affect model output directly. FedTiny does $\Delta R = 10$ rounds of fine-tuning between two finer pruning. When FedTiny reaches $R_{stop}=100$ rounds, it stops pruning and continues fine-tuning.
FedTiny is implemented upon FedML~\cite{he2020fedml}, an open-source machine learning platform that enables lightweight, cross-platform, and provably secure federated learning. 

\begin{figure*}[!tb]
\centering
\includegraphics[width=0.85\textwidth]{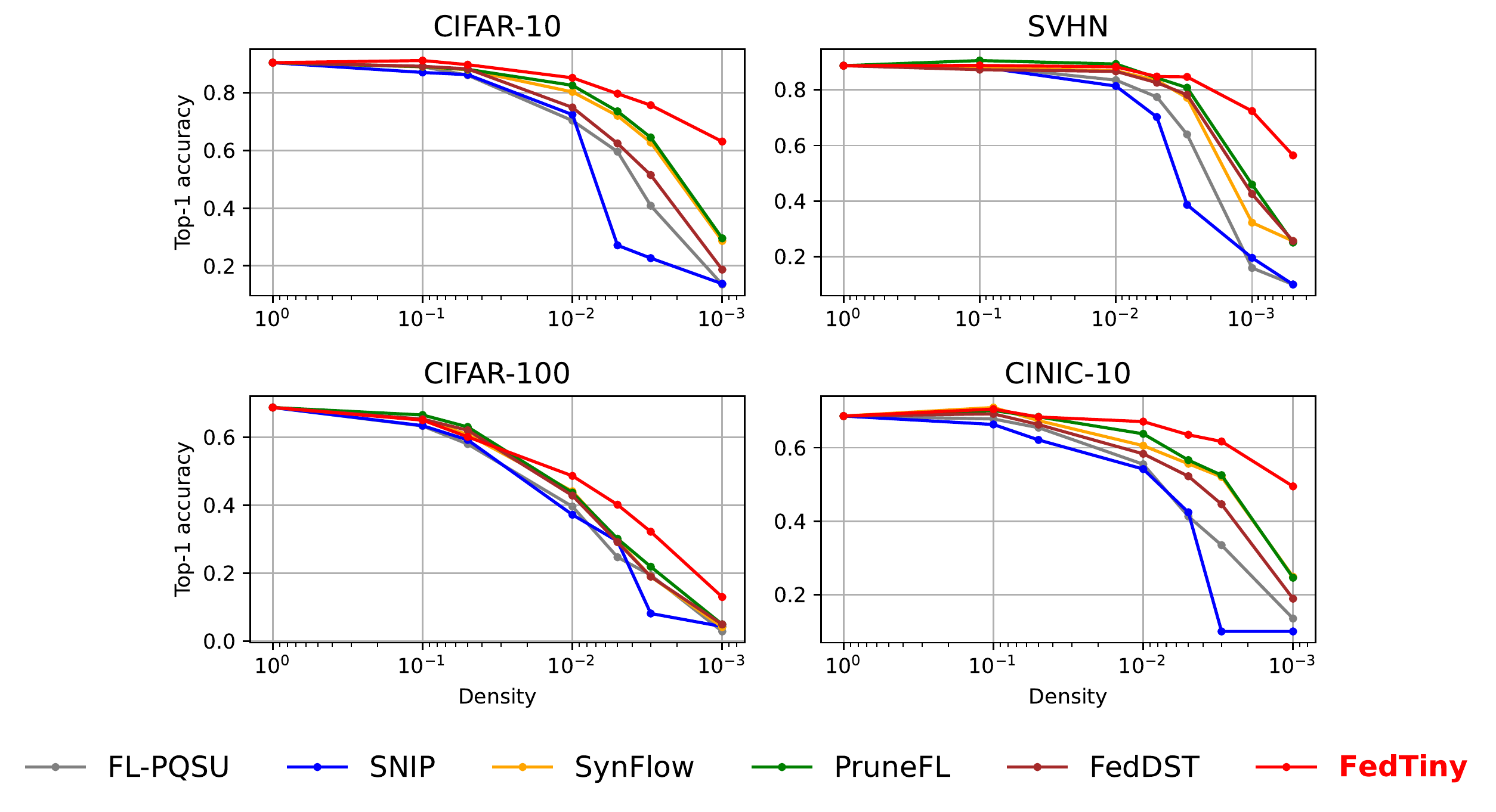} 
\vspace{-1em}
\caption{Top-1 accuracy of different pruning approaches in federated learning. We compare the proposed FedTiny with baselines on the four datasets with different densities. FedTiny outperforms the baselines, especially in the extremely low-density regimes ($<10^{-2}$).}
\label{fig:result}
\end{figure*}

\subsubsection{Baseline Setting}
We involve the following baseline approaches in the study. We include SNIP, SynFlow, and FL-PQSU to confirm that pruning at initialization is not the optimal design choice when the local data are invisible.\footnote{We implement these frameworks based on their open-source implementations \url{https://github.com/ganguli-lab/Synaptic-Flow}.We choose well-known baselines, PruneFL~\cite{jiang2022model}, FedDST~\cite{bibikar2022federated} and LotteryFL~\cite{li2021lotteryfl}, to compare the performance of the FedTiny.} We exclude the FL pruning approaches that are infeasible for memory-constrained FL. For example, FedPrune~\cite{munir2021fedprune} and ZeroFL~\cite{qiu2022zerofl} require powerful devices to continuously process the dense models.
\begin{itemize}
    \item \textbf{SNIP}~\cite{lee2018snip} prunes model by connection sensitivity at initialization with a small public dataset on the server.
    \item \textbf{SynFlow}~\cite{tanaka2020pruning} prunes model by iteratively conserving synaptic flow on the server before training.
    \item \textbf{FL-PQSU}~\cite{xu2021accelerating} prunes model in a one-shot manner based on $l_1$-norm on the server before training. FL-PQSU also includes quantization and selective update parts, but we only use the pruning part in FL-PQSU.
    \item \textbf{PruneFL}~\cite{jiang2022model} uses a powerful device to initially prune the model and applies finer pruning (adaptive pruning) on the sparse model based on full-size averaged gradients. But all devices are resource-constrained in our setting. Therefore, we let PruneFL get the initial pruned model on the server with a small public dataset.
    \item \textbf{LotteryFL}~\cite{li2021lotteryfl} iteratively prunes dense model with a fixed pruning rate on devices and re-initializes the pruned model with the initial values.
    \item \textbf{FedDST}~\cite{bibikar2022federated} first random prunes an initial pruned model on the server, then it deploys the mask adjustment on the devices, and the server uses sparse aggregation and magnitude pruning to obtain a new global model.
\end{itemize}
Since SNIP~\cite{lee2018snip} and PruneFL~\cite{jiang2022model} require some data for coarse pruning, we assume that the server provides a public one-shot dataset $\mathcal{D}_s$ for pretraining. All baselines start with a model pre-trained with the one-shot dataset $\mathcal{D}_s$ on the server.
For SNIP, we apply iterative pruning instead of one-shot pruning as \cite{tanaka2020pruning} shown. Similarly, we let SynFlow prune the model at initialization to the target density in an iterative manner. 
For SNIP and SynFlow, we set 100 pruning epochs on the server at initialization; refer to \cite{tanaka2020pruning}.
For FL-PQSU, which is originally structured pruning, we change it to unstructured pruning since all the other baselines are unstructured pruning frameworks. 
LotteryFL~\cite{li2021lotteryfl} is designed for personalized federated learning, so the model structures are different among devices. Since we attempt to find an optimal structure for all devices as in Eq.~\ref{eq:obj}, we let LotteryFL iteratively prune the global model instead of on-device models to ensure the same model structure for each device.
FedDST~\cite{bibikar2022federated} deploys mask adjustment on devices and fine-tunes the parameters before uploading. We let FedDST adjust the masks after 3 epochs of local training, followed by 2 epochs of fine-tuning.
Since LotteryFL, PruneFL, FedDST, and our FedTiny are iteratively pruning during training, we use the same pruning schedule for these frameworks, where the framework does $\Delta R = 10$ rounds of fine-tuning between two finer pruning. And framework stops pruning and continues fine-tuning after $R_{stop} = 100$ rounds. For PruneFL and FedDST, we set the pruning number $a^l_t$ to be the same as in FedTiny. All baselines will apply uniform sparsity distribution for layer-wise pruning rate setting. 


\begin{table*}[!tb]
\renewcommand\arraystretch{1.2} 
\centering
\caption{Top-1 accuracy and training cost of proposed FedTiny and other baselines with various densities and models}
\resizebox{0.75\linewidth}{!}{
\begin{tabular}{cc|cccccc}
\toprule
\multicolumn{1}{c}{\multirow{3}{*}{Density}} &
  {\multirow{3}{*}{Method}} &
  \multicolumn{3}{c|}{ResNet18} &
  \multicolumn{3}{c}{VGG11} \\ 
\multicolumn{2}{c|}{} &
  \multicolumn{1}{c}{\begin{tabular}[c]{@{}c@{}}Top-1  \\ Accuracy\end{tabular}} &
  \multicolumn{1}{c}{\begin{tabular}[c]{@{}c@{}}Max Training\\  FLOPs \end{tabular}} &
  \multicolumn{1}{c|}{\begin{tabular}[c]{@{}c@{}}Memory \\ Footprint\end{tabular}} &
  \multicolumn{1}{c}{\begin{tabular}[c]{@{}c@{}}Top-1  \\ Accuracy\end{tabular}} &
  \multicolumn{1}{c}{\begin{tabular}[c]{@{}c@{}}Max Training\\  FLOPs \end{tabular}} &
  \multicolumn{1}{c}{\begin{tabular}[c]{@{}c@{}}Memory \\ Footprint\end{tabular}} \\ \midrule
\multicolumn{1}{c}{1} &
  FedAvg &
  0.9048 &
  1x(8.33E13) &
  \multicolumn{1}{c|}{90.91MB} &
  0.8696 &
  1x(4.09E13) &
  \multicolumn{1}{c}{1033.33MB} \\ \hline
\multicolumn{1}{c}{\multirow{6}{*}{0.01}} &
  FL-PQSU &
  0.7038 &
  0.014x &
  \multicolumn{1}{c|}{2.75MB} &
  0.475 &
  0.017x &
  \multicolumn{1}{c}{20.96MB} \\ 
\multicolumn{1}{c}{} &
  SNIP &
  0.7245 &
  0.014x &
  \multicolumn{1}{c|}{2.76MB} &
  0.3481 &
  0.017x &
  \multicolumn{1}{c}{20.98MB} \\ 
\multicolumn{1}{c}{} &
  SynFlow &
  0.8034 &
  0.014x &
  \multicolumn{1}{c|}{2.75MB} &
  0.5803 &
  0.017x &
  \multicolumn{1}{c}{20.92MB}  \\ 
\multicolumn{1}{c}{} &
  PruneFL &
  \textbf{\textcolor[rgb]{0,0.1,1}{0.8262}} &
  0.34x &
  \multicolumn{1}{c|}{46.58MB} &
  \textbf{\textcolor[rgb]{0,0.1,1}{0.6204}} &
  0.34x &
  \multicolumn{1}{c}{526.87MB} \\ 
\multicolumn{1}{c}{} &
  FedDST &
  0.7495 &
  0.015x &
  \multicolumn{1}{c|}{2.91MB} &
  0.6067 &
  0.017x &
  \multicolumn{1}{c}{21.03MB} \\ 
\multicolumn{1}{c}{} &
  LotteryFL &
  0.8083 &
  1x &
  \multicolumn{1}{c|}{90.91MB} &
  0.6183 &
  1x &
  \multicolumn{1}{c}{1033.33MB} \\ 
\multicolumn{1}{c}{} &
  \textbf{FedTiny} &
  \textbf{\textcolor[rgb]{1,0.25,0}{0.8523}} &
  0.014x &
  \multicolumn{1}{c|}{2.79MB} &
  \textbf{\textcolor[rgb]{1,0.25,0}{0.7883}} &
  0.017x &
  \multicolumn{1}{c}{20.95MB} \\ \hline
\multicolumn{1}{c}{\multirow{6}{*}{0.005}} &
  FL-PQSU &
  0.5961 &
  0.008x &
  \multicolumn{1}{c|}{2.01MB} &
  0.232 &
  0.012x &
  \multicolumn{1}{c}{11.99MB} \\ 
\multicolumn{1}{c}{} &
  SNIP &
  0.2711 &
  0.009x &
  \multicolumn{1}{c|}{1.98MB} &
  0.2409 &
  0.012x &
  \multicolumn{1}{c}{12.00MB} \\ 
\multicolumn{1}{c}{} &
  SynFlow &
  0.7206 &
  0.008x &
  \multicolumn{1}{c|}{2.00MB} &
  0.4376 &
  0.012x &
  \multicolumn{1}{c}{11.95MB} \\ 
\multicolumn{1}{c}{} &
  PruneFL &
  0.736 &
  0.34x &
  \multicolumn{1}{c|}{46.19MB} &
  \textbf{\textcolor[rgb]{0,0.1,1}{0.4956}} &
  0.34x &
  \multicolumn{1}{c}{522.31MB} \\ 
\multicolumn{1}{c}{} &
  FedDST &
  0.6245 &
  0.009x &
  \multicolumn{1}{c|}{2.04MB} &
  0.4292 &
  0.013x &
  \multicolumn{1}{c}{12.02MB} \\ 
\multicolumn{1}{c}{} &
  LotteryFL &
  \textbf{\textcolor[rgb]{0,0.1,1}{0.7586}} &
  1x &
  \multicolumn{1}{c|}{90.91MB} &
  0.4376 &
  1x &
  \multicolumn{1}{c}{1033.33MB} \\ 
\multicolumn{1}{c}{} &
  \textbf{FedTiny} &
  \textbf{\textcolor[rgb]{1,0.25,0}{0.7972}} &
  0.009x &
  \multicolumn{1}{c|}{2.03MB} &
  \textbf{\textcolor[rgb]{1,0.25,0}{0.7534}} &
  0.012x &
  \multicolumn{1}{c}{11.98MB} \\ \hline
  
  \multicolumn{1}{c}{\multirow{6}{*}{0.001}} &
  FL-PQSU &
  0.1352 &
  0.004x &
  \multicolumn{1}{c|}{1.22MB} &
  0.1 &
  0.008x &
  \multicolumn{1}{c}{4.71MB} \\ 
\multicolumn{1}{c}{} &
  SNIP &
  0.1377 &
  0.004x &
  \multicolumn{1}{c|}{1.19MB} &
  0.1 &
  0.008x &
  \multicolumn{1}{c}{4.72MB} \\ 
\multicolumn{1}{c}{} &
  SynFlow &
  0.2862 &
  0.004x &
  \multicolumn{1}{c|}{1.19MB} &
  0.2531 &
  0.008x &
  \multicolumn{1}{c}{4.72MB}  \\ 
\multicolumn{1}{c}{} &
  PruneFL &
  0.2955 &
  0.336x&
  \multicolumn{1}{c|}{45.72MB} &
  \textbf{\textcolor[rgb]{0,0.1,1}{0.2692}} &
  0.339x &
  \multicolumn{1}{c}{518.71MB} \\ 
\multicolumn{1}{c}{} &
  FedDST &
  0.1868 &
  0.004x &
  \multicolumn{1}{c|}{1.20MB} &
  0.2535 &
  0.008x &
  \multicolumn{1}{c}{4.73MB} \\ 
\multicolumn{1}{c}{} &
  LotteryFL &
  \textbf{\textcolor[rgb]{0,0.1,1}{0.3070}} &
  1x  &
  \multicolumn{1}{c|}{90.91MB} &
  0.2634 &
  1x &
  \multicolumn{1}{c}{1033.33MB} \\ 
\multicolumn{1}{c}{} &
  \textbf{FedTiny} &
  \textbf{\textcolor[rgb]{1,0.25,0}{0.6311}} &
  0.004x &
  \multicolumn{1}{c|}{1.17MB} &
  \textbf{\textcolor[rgb]{1,0.25,0}{0.5944}} &
  0.008x &
  \multicolumn{1}{c}{4.71MB} \\ \bottomrule
\end{tabular}
}

\label{tb:baseline}
\end{table*}

\subsection{Comparison Between FedTiny and Baseline Approaches} 
In order to show the performance of FedTiny under different densities, we compare baselines and FedTiny on four datasets (CIFAR-10, CIFAR-100, CINIC-10, and SVHN) with ResNet18. As shown in Fig.~\ref{fig:result}, FedTiny outperforms the other baselines in the low-density regime ($d_{target} < 10^{-2}$), \textit{e.g.,} FedTiny achieves an accuracy improvement of 18.91\% in SVHN dataset compared to state-of-the-art methods with $10^{-3}$ density. This benefits from the adaptive batch normalization selection module used in FedTiny, which deduces an adaptive coarse pruning structure on the server. This initial pruning structure has less bias, reducing the size of the search space and improving convergence.
Besides, FedTiny is competitive with a high density ($d_{target} >10^{-2}$), \textit{e.g.,} FedTiny outperforms state-of-the-art methods with $10^{-1}$ density by 1.3\% in CIFAR-10 dataset. Although PruneFL can partially outperform FedTiny under high density, it requires over $20\times$ computation cost and $15\times$ memory footprint to process dense importance scores on devices. SNIP performs badly in low density because SNIP tends to remove nearly all parameters in some layers. Moreover, the pruned model in SNIP highly depends on the samples on the server, which increases bias due to non-iid. We do not include LotteryFL in Fig.~\ref{fig:result} as the utilization of LotteryFL necessitates the training of a large model, which incurs a substantial computational cost and memory footprint. However, the results of LotteryFL are included in Table~\ref{tb:baseline} for the purpose of providing a comprehensive comparison with other baselines.

To show the efficiency of FedTiny, we measure the cost of training ResNet18 and VGG11 with various densities on the CIFAR-10 dataset.  
We use the number of floating point operations (FLOPs) to measure the computational cost for each device. \hong{The pruning operation requires a variable amount of computation per round, resulting in variable training FLOPs per round. Therefore, we report the maximum training FLOPs per round (Max Training FLOPs).The maximum training FLOPs per round is used to evaluate whether devices suffer from intensive computation in a single round. We also report the memory footprint in devices, which is related to memory cost in deployment. } 

Table~\ref{tb:baseline} shows the accuracy and training cost of proposed FedTiny and other baselines with various densities and models. We mark the best metric in red and the second-best metric in blue. All cost measurements are for one device in one pruning round. We also report the performance of FedAvg to show the upper bound of pruning approaches. As shown in Table~\ref{tb:baseline},  FedTiny aims to improve both accuracy and memory efficiency. The existing works cannot achieve satisfactory accuracy in ultra-low density. Our proposed FedTiny  significantly improves the accuracy with the lowest levels of FLOPs and memory footprint.


\begin{figure}[!tb]
\centering
\includegraphics[width=0.8\columnwidth]{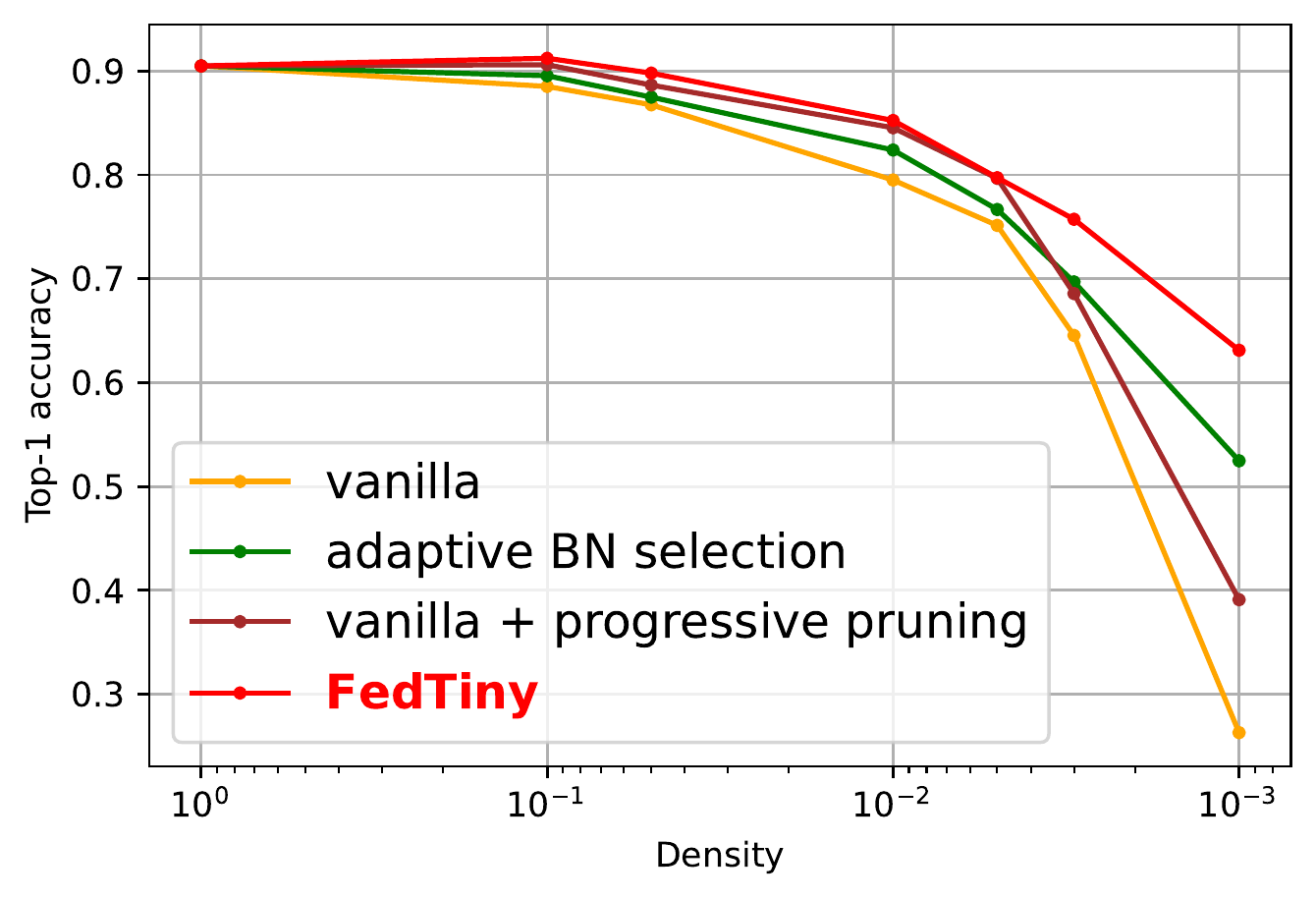} 
\caption{Ablation studies the two key modules in FedTiny: the adaptive batch normalization selection module and the progressive pruning module. We compare vanilla selection, adaptive batch normalization (BN) selection, progressive pruning with vanilla selection, and FedTiny. We test the ResNet18 model on the CIFAR-10 dataset with various densities.}
\label{fig:ablation}
\vspace{-1em}
\end{figure}

\begin{figure*}
	\begin{minipage}[]{0.5\linewidth}
		\centering
		\includegraphics[width=0.7\columnwidth]{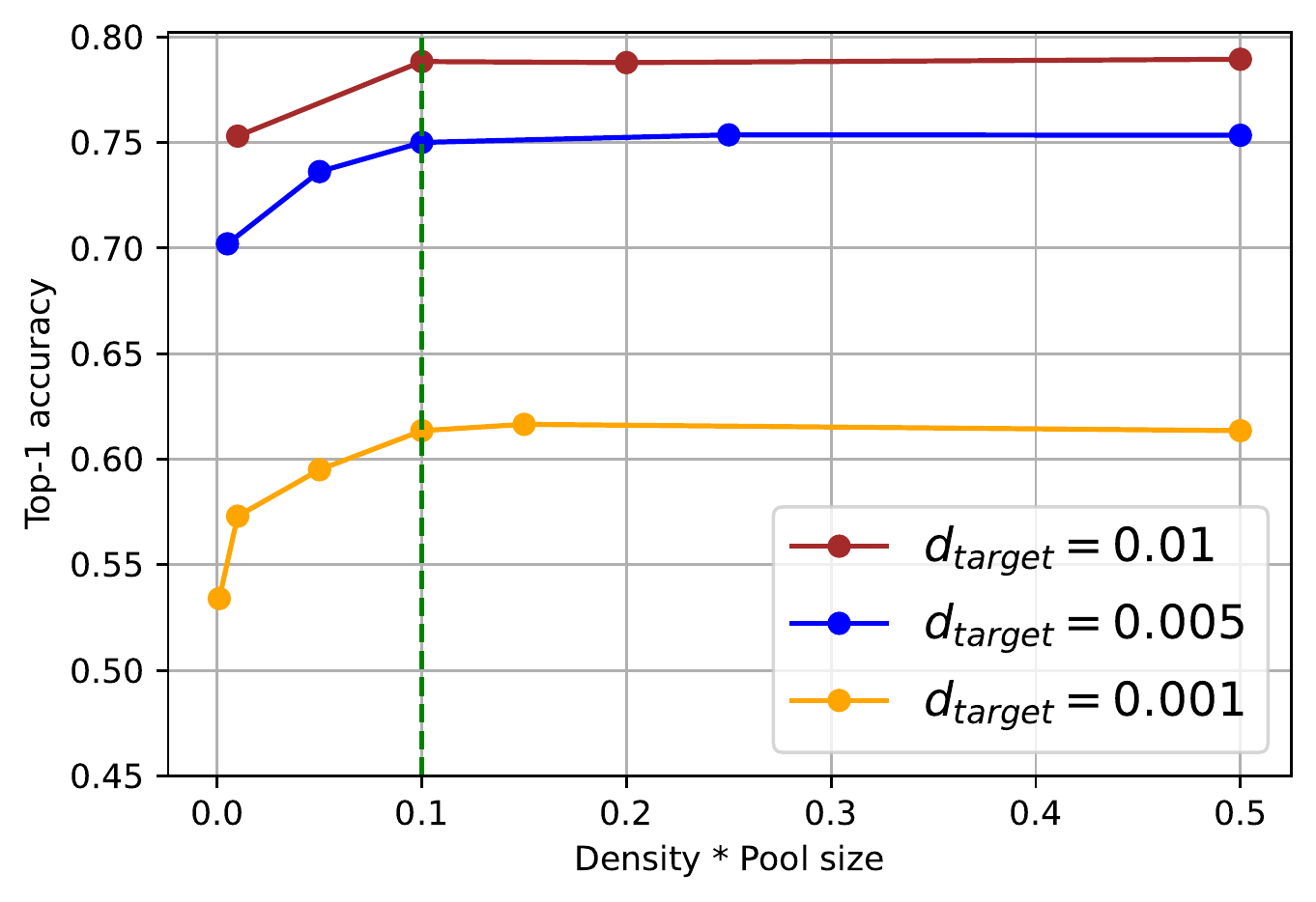}
	\end{minipage}
	\begin{minipage}[]{0.5\linewidth}
		\centering
		\includegraphics[width=0.7\columnwidth]{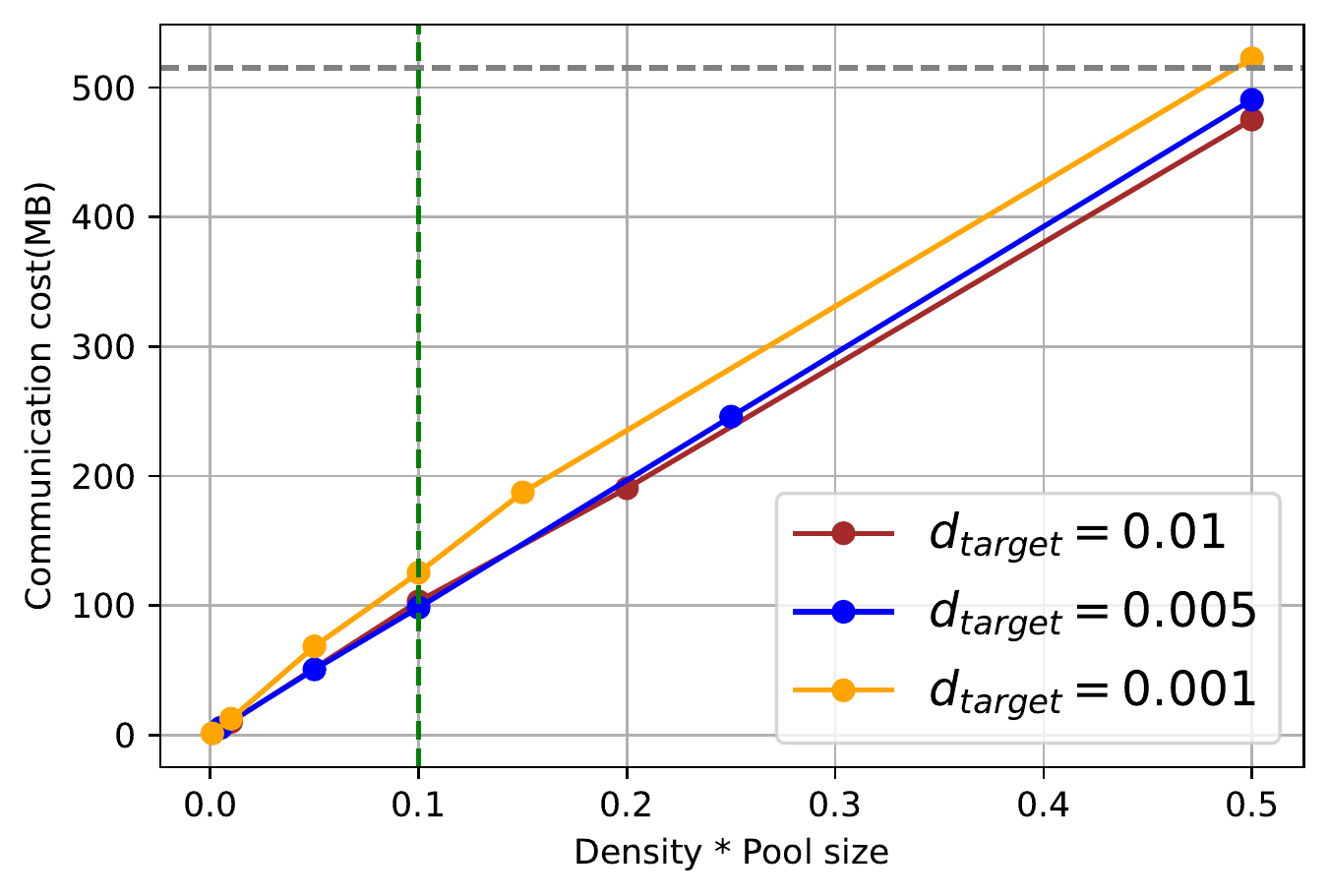}
	\end{minipage}
	\caption{ The performance and cost for sparse VGG11 models with different densities and pool sizes. \textit{Left}: The effect of pool size on top-1 accuracy under different densities. \textit{Right}: The effect of pool size on communication cost in the adaptive batch normalization selection module under different densities. The gray dash line is the size for a full-size VGG11 model. The green dash line is the optimal candidate pool size for specific density.}
	\label{fig:pool_size}
\end{figure*}

\subsection{Ablation Study}
This section discusses the effectiveness of each module in FedTiny via ablation studies.
We evaluate vanilla selection, adaptive batch normalization selection, progressive pruning after vanilla selection, and FedTiny on the CIFAR-10 dataset with the VGG11 model. 
Fig.~\ref{fig:ablation} shows the results of each module working individually. 
We have the following three findings. 
First, both the adaptive batch normalization selection module and progressive pruning module improve the performance in vanilla selection, indicating the effectiveness of these two modules.
Second, a coarse-pruned model from adaptive batch normalization selection faces a drop in accuracy compared to FedTiny, indicating that there are still some biases in the selected coarse-pruned model, and the progressive pruning module can remove them. 
Last, the progressive pruning module with vanilla selection reaches the same level of accuracy compared to FedTiny with the high density ($<10^{-2}$). However, it suffers from severe degradation of accuracy in the low-density regime ($>10^{-2}$), which suggests that the progressive pruning module only removes the bias to a certain extent, and it must be combined with the adaptive batch normalization selection module in the low-density regime.
Therefore, independently using the adaptive batch normalization selection module and progressive pruning module can improve performance, but the improvement is limited. The combination of the two modules, \textit{i.e.}, FedTiny,  achieves the best prediction performance with the tiny model.

\subsection{Overhead in Adaptive BN Selection Module}
\label{sec:eval_impact}
Although a larger candidate pool provides more choices for selection, it brings more communication costs in the adaptive batch normalization selection module. So, we want to find an optimal pool size that can trade off the accuracy and overhead in the adaptive batch normalization selection module. Therefore, We evaluate FedTiny on VGG11 with different pool sizes to find an optimal pool size.
We do the experiments on CIFAR-10 datasets with the VGG11 model with different pool sizes and densities. The experiment results are shown in Fig.~\ref{fig:pool_size}.
The result shows increasing the pool size beyond the green line may only yield a marginal increase in accuracy, while significantly increasing computational costs. Therefore, the green line serves as a practical threshold for selecting the optimal candidate pool size.
Therefore, the optimal pool size is selected as $C^* = \frac{0.1}{d_{target}}$ for specific density $d_{target}$, where the communication cost in adaptive batch normalization selection module is as low as $20\%$ to a full-size VGG11 model, and FedTiny can receive a relatively good accuracy. A larger pool size $>C^*$ slightly improves accuracy but incurs much higher communication costs. 

We also calculate the extra FLOPs for the adaptive batch normalization selection module with optimal pool size, as shown in Table~\ref{tab:pool_flop}. The extra FLOPs in adaptive batch normalization selection are less than one round of sparse training. Since federated learning usually involves more than one hundred rounds of training, the extra computational overhead is neglectable. Therefore, we argue that the overhead introduced by adaptive batch normalization selection is marginal.

\begin{table}[!tb]
\centering
\caption{Extra FLOPs in the adaptive BN selection model}
\label{tab:pool_flop}
\begin{tabular}{cccc}
\hline
Density & Pool Size & \multicolumn{1}{c}{\begin{tabular}[c]{@{}c@{}}Extra FLOPs\\ in selection\end{tabular}} & \begin{tabular}[c]{@{}c@{}}Training FLOPs\\ in one round\end{tabular} \\ \hline
0.01    & 10        & 9.15E+10                                                                               & 6.86E+11                                                              \\
0.005   & 20        & 1.3E+11                                                                                & 4.92E+11                                                              \\
0.001   & 100       & 3.42E+11                            & 3.56E+11                                                              \\ \hline
\end{tabular}
\end{table}

\begin{table}[!tb]
\centering
\caption{Top-1 accuracy for FedTiny with different pruning scheduling strategies}
\begin{tabular}{cc|ccc}
\toprule
Granularity & $\Delta R$/$R_{stop}$ & \begin{tabular}[c]{@{}c@{}}Density\\  0.01\end{tabular} & \begin{tabular}[c]{@{}c@{}}Density\\  0.005\end{tabular} & \begin{tabular}[c]{@{}c@{}}Density\\  0.001\end{tabular}\\ \midrule
Layer           & 5/100 & \multicolumn{1}{c}{0.7623} & \multicolumn{1}{c}{0.7034} & 0.447  \\ 
Layer ($b$) & 5/100 & \multicolumn{1}{c}{\textcolor{red}{\textbf{0.7894}}} & \multicolumn{1}{c}{0.7343} & 0.5871 \\ 
Block           & 10/100 & \multicolumn{1}{c}{0.7697} & \multicolumn{1}{c}{0.7179} & 0.5721 \\ 
Block ($b$) & 10/100  & \multicolumn{1}{c}{\textcolor{blue}{\textbf{0.7883}}} & \multicolumn{1}{c}{\textcolor{red}{\textbf{0.7534}}} & \textcolor{red}{\textbf{0.6311}} \\ 
Block ($b$) & 5/50   & \multicolumn{1}{c}{0.7675} & \multicolumn{1}{c}{0.7263} & 0.6113 \\ 
Entire          & 50/100 & \multicolumn{1}{c}{0.772}  & \multicolumn{1}{c}{\textcolor{blue}{\textbf{0.7395}}} & \textcolor{blue}{\textbf{0.6244}} \\ 
Entire          & 25/50   & \multicolumn{1}{c}{0.7583} & \multicolumn{1}{c}{0.7043} & 0.5944 \\ \bottomrule
\end{tabular}
\label{tb:grand}
\end{table}

\subsection{Impact of Pruning Scheduling Strategy}
\label{sec:impact_sch}
Although layer-wise adjustment in progressive pruning reduces the computation cost in one round, it may slow down the convergence speed. To determine the best pruning granularity and pruning frequency, we evaluate FedTiny on VGG11 with different pruning granularities (one layer per round, one block per round, and the entire model per round) and different pruning frequencies. 

Table~\ref{tb:grand} shows the top-1 accuracy of various pruning schedules under different densities on VGG11 with the CIFAR-10 dataset, where the best performance of Top-1 accuracy with the same density is represented in red, and the second-best metric is marked in blue. $b$ denotes sequentially choosing layers or blocks to prune in backward order, \textit{i.e.,} from the output layer to the input layer. We control the pruning frequencies by setting different interval rounds $\Delta R$ between two pruning. If the pruning granularity is too small (\textit{e.g.}, layer-wise pruning), the model structure will converge slowly, and the optimal structure cannot be achieved with limited training resources. But high updating granularity leads to more intensive computation in one round. We find that pruning a block per round is an optimal choice for the progressive pruning module.
Moreover, sequentially choosing blocks to prune in backward order (from the output layer to the input layer) gets better results than forwarding order since the gradient propagation is backward, and we use gradients to adjust the model structure.

\subsection{Effectiveness of FedTiny Over Heterogeneous Data Distributions}
\label{sec:non_iid}
Neural network pruning requires training data to determine the proper model structure. Due to resource-constrained devices, the server cannot push the dense model to devices. Therefore, the server needs to coarsely prune to produce the initial pruned model. Due to privacy concerns in federated learning, the server cannot know the data distributions for all devices. In the existing methods, the server only coarsely prunes the model based on the pre-trained dataset or the data from some trusted devices. It makes the dataset used for pruning different from the dataset used for fine-tuning, which causes bias in coarse pruning. Therefore, our strategy is to use adaptive batch normalization selection to select one pruned model with less bias. 

\begin{figure}[!tb]
\centering
\includegraphics[width=0.7\columnwidth]{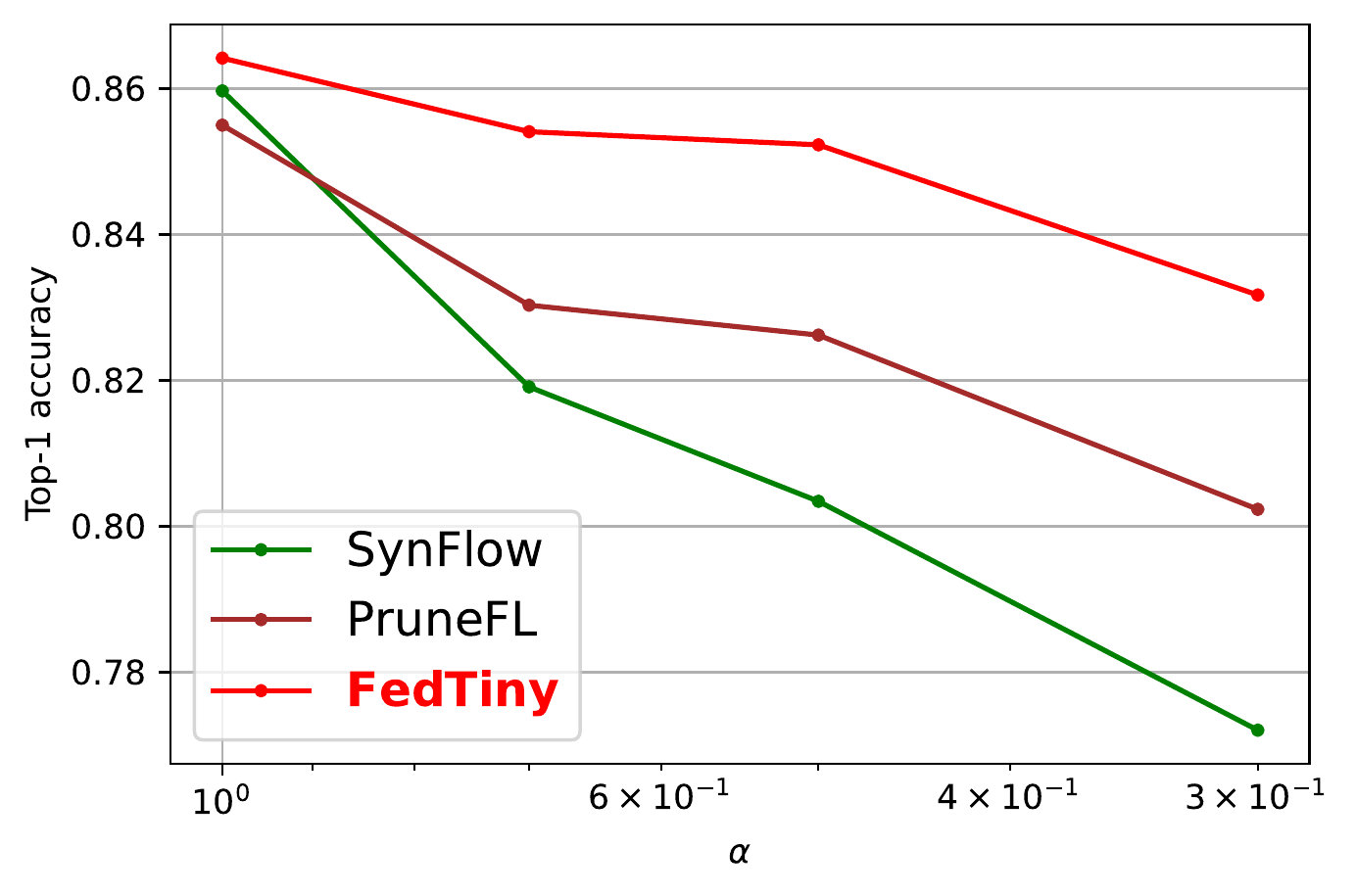} 
\caption{Top-1 accuracy of different pruning approaches on various non-iid degrees. Lower $\alpha$ indicates a higher non-iid degree.}
\label{fig:noniid}
\vspace{-1em}
\end{figure}

To demonstrate the effectiveness of FedTiny over heterogeneous data distributions, we set different non-iid degrees by using different $\alpha$ in the Dirichlet distribution. Lower $\alpha$ indicates a higher non-iid degree. We do experiments on the CIFAR-10 dataset with ResNet18 with 1\% density. The experiments are shown in Fig.~\ref{fig:noniid}. Our experiments show that 1) the performance of the existing pruning methods (\textit{e.g.}, SynFlow, PruneFL) in Federated Learning will be significantly degraded given a higher non-iid degree; 2) Our proposed FedTiny mitigates the bias in pruning and achieves the best performance compared with the existing pruning methods.

\subsection{Comparison Between FedTiny And Small Model Training}
In the previous experiments, we compare the FedTiny with existing pruning methods. To further investigate the effectiveness of FedTiny, we compare FedTiny with the dense small models without pruning. FedTiny outperforms other baselines in the very sparse model, like 1\% density. In this case, training a dense small model without pruning can also be considered as a baseline. Therefore, we design the experiments on small models. We train a small model with three convolutional layers. First, we evaluate the small model with a similar number of parameters to ResNet18 with 1\% density on different datasets. Second, we evaluate the small network with a similar number of parameters to ResNet18 with different densities on CIFAR-10.
 We also choose SynFlow and PruneFL as references. The experiment result is shown in Table~\ref{tab:smallcnn1} and Table~\ref{tab:smallcnn2}, where the best performance of Top-1 accuracy with the same dataset and same density is represented in red, and the second-best metric is marked in blue. The experimental results show that the small model is competitive compared to other baselines. However, our proposed FedTiny achieves much better performance than the small model, which demonstrates the advantage of FedTiny.

\begin{table}[]
\centering
\caption{Top-1 accuracy for ResNet18 with 1\% density and a small model}
\begin{tabular}{c|cccc}
\hline
Method   & CIFAR-10 & CINIC-10 & SVHN & CIFAR-100 \\ \hline
SynFlow & 0.8034  & 0.6057    & 0.8683 & \textcolor{blue}{\textbf{0.4413}}  \\
PruneFL  & \textcolor{blue}{\textbf{0.8262}}  & \textcolor{blue}{\textbf{0.6379}}    & \textcolor{red}{\textbf{0.8927}} &  0.4373  \\
Small Model & 0.8019  & 0.5578    & 0.8395 & 0.4277  \\
FedTiny & \textcolor{red}{\textbf{0.8523}}  & \textcolor{red}{\textbf{0.6712}}    & \textcolor{blue}{\textbf{0.8826}} & \textcolor{red}{\textbf{0.4865}}  \\ \hline
\end{tabular}
\label{tab:smallcnn1}
\end{table}

\begin{table}[]
\centering
\caption{Top-1 accuracy for ResNet18 with various densities and small models on the CIFAR-10 dataset}
\begin{tabular}{c|cccc}
\hline
Method & 0.01 & 0.005 & 0.003   & 0.001 \\ \hline
SynFlow    & 0.8034   & 0.7206    & 0.6279 & 0.2862  \\
PruneFL  & \textcolor{blue}{\textbf{0.8262}}  & \textcolor{blue}{\textbf{0.7360}}     & 0.6453 & 0.2955 \\
Small Model   & 0.8019  & 0.7201    & \textcolor{blue}{\textbf{0.6921}} & \textcolor{blue}{\textbf{0.6158}}  \\
FedTiny   & \textcolor{red}{\textbf{0.8523}}  & \textcolor{red}{\textbf{0.7972}}    & \textcolor{red}{\textbf{0.7572}} & \textcolor{red}{\textbf{0.6311}}  \\ \hline
\end{tabular}

\label{tab:smallcnn2}
\end{table}
\section{Conclusion}

This paper develops a novel distributed pruning framework called FedTiny.
FedTiny enables memory-efficient local training and determines specialized tiny models in federated learning for different deployment scenarios (participating hardware platforms and training tasks).
FedTiny addresses the challenges of bias, intensive computation, and memory usage that existing federated pruning research suffers.
FedTiny introduces two critical modules: an adaptive batch normalization selection module and a lightweight progressive pruning module. The batch normalization selection module is designed to mitigate the bias in pruning caused by the heterogeneity of local data, while the progressive pruning module enables fine-grained pruning under strict computational and memory budgets. Specifically, it gradually determines the pruning policy for each layer rather than evaluating the overall model structure.
Experimental results demonstrate the effectiveness of FedTiny when compared to state-of-the-art approaches. In particular, FedTiny achieves significant improvements in terms of accuracy, FLOPs, and memory footprint when compressing deep models to extremely sparse tiny models. The results on the CIFAR-10 dataset show that FedTiny outperforms state-of-the-art methods by achieving an accuracy improvement of 2.61\% while simultaneously reducing FLOPs by 95.9\% and memory footprint by 94.0\%. The experimental results demonstrate the effectiveness and efficiency of FedTiny in federated learning settings.
\section{Acknowledgement}
This work is supported in part by the National Science Foundation (CCF-2221741, CCF-2106754, CNS-2151238, CNS-2153381), the ORAU Ralph E. Powe Junior Faculty Enhancement Award, and the Hong Kong Research Grants Council, General Research Fund (GRF) under Grant 11203523.

\bibliographystyle{ieeetr}
\bibliography{main}


\end{document}